\documentclass[letterpaper]{article} 
\usepackage{aaai2026}  
\usepackage{times}  
\usepackage{helvet}  
\usepackage{courier}  
\usepackage[hyphens]{url}  
\usepackage{graphicx} 
\usepackage{natbib}  
\usepackage{caption} 
\usepackage{booktabs}
\usepackage{amsmath} 
\usepackage{algorithm}
\usepackage{algorithmic}
\usepackage[table,dvipsnames]{xcolor} 
\usepackage{enumitem}
\usepackage{newfloat}
\usepackage{listings}
\DeclareCaptionStyle{ruled}{labelfont=normalfont,labelsep=colon,strut=off} 
\floatstyle{ruled}
\newfloat{listing}{tb}{lst}{}
\floatname{listing}{Listing}
\usepackage{amsmath}
\usepackage{float}
\usepackage{placeins} 
\usepackage{tikz}

\newcommand{\topone}[1]{\cellcolor{gray!60}#1}
\newcommand{\toptwo}[1]{\cellcolor{gray!40}#1}
\newcommand{\topthree}[1]{\cellcolor{gray!20}#1}
\usepackage[most]{tcolorbox}
\usepackage{lipsum}

\title{Can Structured Templates Facilitate LLMs in Tackling Harder Tasks? : An Exploration of Scaling Laws by Difficulty}
\author {
    Zhichao Yang\textsuperscript{\rm 1,\rm 2},
    Zhaoxin Fan\textsuperscript{\rm 1,\rm 2},
    Gen Li\textsuperscript{\rm 1,\rm 2},
    Yuanze Hu\textsuperscript{\rm 1,\rm 2},
    Xinyu Wang\textsuperscript{\rm 1,\rm 2},
    Ye Qiu\textsuperscript{\rm 1,\rm 2},
    Xin Wang\textsuperscript{\rm 1},
    Yifan Sun\textsuperscript{\rm 1,\rm 2},
    Wenjun Wu\textsuperscript{\rm 1,\rm 2}
}
\affiliations{
    \textsuperscript{\rm 1}School of Artificial Intelligence, Beihang University, Beijing, China\\
    \textsuperscript{\rm 2}High-precision and Cutting-edge Innovation Center for Future Blockchain and Privacy Computing, Beijing, China\\
    yzc0912@buaa.edu.cn
}

\usepackage{bibentry}

\begin{document}
\maketitle

\begin{abstract}

Structured, procedural reasoning is essential for Large Language Models (LLMs), especially in mathematics. While post-training methods have improved LLM performance, they still fall short in capturing deep procedural logic on complex tasks. To tackle the issue, in this paper, we first investigate this limitation and uncover a novel finding: a Scaling Law by Difficulty, which reveals that model performance follows a U-shaped curve with respect to training data complexity—excessive low-difficulty data impedes abstraction, while high-difficulty data significantly enhances reasoning ability. Motivated by this, we propose the Structured Solution Template (SST) framework, which uses solution templates and a curriculum of varied difficulty to explicitly teach procedural reasoning. Specifically, SST comprises (1) fine-tuning with structured solution-template chains and dynamically weighted loss to prioritize procedural logic, (2) prompt-time injection of solution templates as cognitive scaffolds to guide inference, and (3) integrated curriculum fine-tuning that explicitly teaches the model to self-plan → execute → self-correct.
 Experiments on GSM8K, AIME24, and new Dynamic En benchmark show that SST significantly improves both accuracy and efficiency, especially on harder problems.

\end{abstract}


\section{Introduction}



Large language models (LLMs) such as DeepSeek-R1~\cite{guo2025deepseek}, OpenAI-o1~\cite{jaech2024openai}, and Qwen~\cite{yang2025qwen3} have driven remarkable progress across natural language understanding, dialogue generation, and code synthesis. Their ability to produce contextually coherent text has unlocked powerful applications in content creation, question answering, and, increasingly, mathematical problem solving. As LLMs become integral to scientific research, education, and engineering, robust reasoning, especially in mathematics, is emerging as a crucial benchmark for machine intelligence.

\begin{figure}[!t]  
    \centering 
    \includegraphics[width=\linewidth]{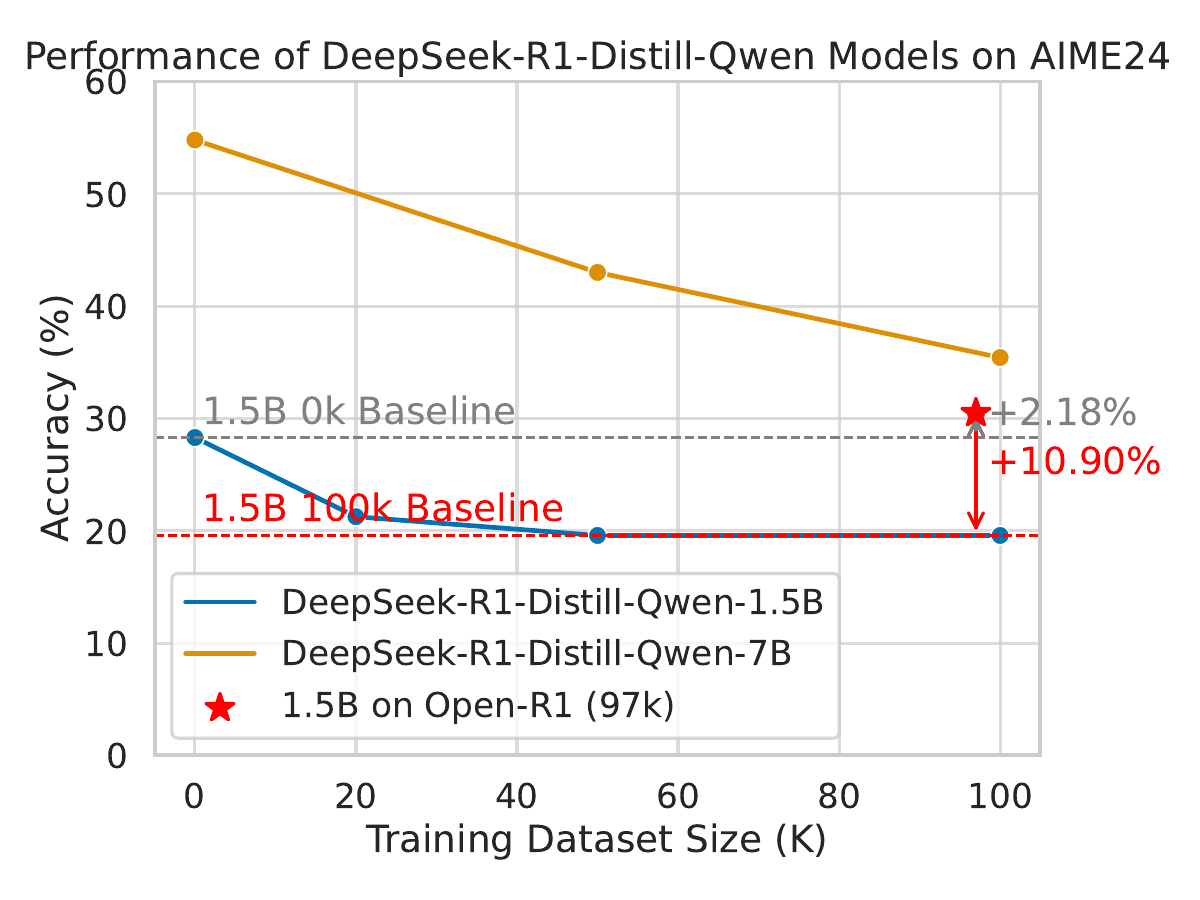}  
    \caption{ \textbf{Illustration of the Scaling Law by Difficulty.} Model performance declines as the amount of template-generated synthetic data increases: as dataset size grows from 0k to 100k, the accuracy of DeepSeek-R1-Distill-Qwen on AIME24 consistently drops, even though the training and test problems are structurally similar. This counter-intuitive result highlights a key limitation—large quantities of low-abstraction data lead to rote pattern imitation rather than true procedural reasoning.}

    \label{fig:template_workflow}
\end{figure}

To enhance LLM reasoning capabilities, numerous post-training techniques have been developed, including chain-of-thought prompting, instruction fine-tuning, and synthetic data generation~\cite{long2024llms,zhao2025promptcot, wei2022chain,wei2021finetuned}. These approaches have led to significant gains on a range of benchmarks~\cite{ahn2024large, zhu2022solving}. However, despite these advances, fundamental limitations persist. LLMs remain next-token predictors trained predominantly on unstructured text, lacking explicit mechanisms for structured knowledge representation, state tracking, and symbolic manipulation~\cite{huang2022towards,zelikman2022star}. As a result, they often rely on surface pattern matching rather than genuine procedural abstraction~\cite{anantheswaran2024investigating,imani2023mathprompter}. This deficiency is particularly pronounced in mathematical and procedural reasoning, where even minor structural changes in a problem can lead to sharp performance drops~\cite{wang2023math, zheng2024processbench}.

In an effort to address these shortcomings, we explore post-training with large amounts of template-generated data, hoping this will help LLMs learn solution methods. However, our results show the opposite: as the dataset size increases, model accuracy on hard math tasks like AIME24 actually drops (see Fig.~\ref{fig:template_workflow}). We call this the Scaling Law by Difficulty: too much simple, repetitive data leads the model to memorize patterns instead of learning real problem-solving skills, which hurts its ability to generalize.




Based on this insight, we design the Structured Solution Template (SST) framework, which turns abstract reasoning into a clear learning goal and helps overcome the problems shown by the Scaling Law by Difficulty.  The SST framework consists of a three-stage data construction and training process. In Stage1, we turn high-quality chain-of-thought solutions into short \verb|<chain>...</chain>| templates and give them extra importance during training, so the model learns to focus on reasoning steps instead of just copying text. In Stage2, we use a lightweight LoRA-based generator to create problem-specific \verb|<chain>| prompts on the fly. These prompts are added to each input to improve reasoning and efficiency, without changing the main model’s weights. In Stage3, we combine both steps into a plan-then-execute process with self-reflection: we collect the most challenging errors from Stage1, add chain templates and self-reflection notes using DeepSeek-R1, and use Group Relative Policy Optimization (GRPO) to help the model learn to plan, solve, and self-correct problems.

To show the effectiveness of SST, we conduct experiments on 7 different benchmarks, including GSM8K, MATH500, AIME 2024/25, AMC 2023, Gaokao English, and our new Dynamic En dataset for cross-domain reasoning. SST outperforms all previous methods, with up to a 6.2 point gain on GSM8K, a 2.2 point increase on AIME 24, and steady improvements on Dynamic Math. The results significantly demonstrate that using difficulty-aware, explicit templates helps LLMs truly learn step-by-step reasoning. Our contributions can be summarized as:
\begin{itemize}
  \item \textbf{Scaling Law by Difficulty:} We identify and validate the U-shaped relationship between training data difficulty and model reasoning performance, showing that mere quantity of low-difficulty synthetic samples cannot cultivate deep abstraction.
  \item \textbf{SST Framework:} We propose a three-stage method—weighted chain supervision, dynamic chain injection, and a unified plan-then-execute mechanism with self-reflection. Furthermore, our SST results validate and extend the Scaling Law by Difficulty by showing that explicit chain scaffolding both prevents performance collapse on easy examples and amplifies gains on the most challenging problems, effectively reshaping the U-shaped difficulty–performance curve.

  \item \textbf{Empirical Gains:} We demonstrate significant improvements across top-tier benchmarks—up to +6.2 pts on GSM8K and +2.2 pts on AIME 24—confirming that template-guided, difficulty-aware training is crucial for advanced LLM reasoning.
\end{itemize}

\section{Related Work}

\textbf{Automated problem generation.}  
Recent efforts use LLMs to synthesize large-scale math data: KPDDS extracts key solution steps into QA pairs, MathScale builds concept graphs for millions of questions~\cite{huang2025key}, and OpenMathInstruct-2 provides 14M instruction–answer pairs~\cite{toshniwal2024openmathinstruct}. Systems like NuminaMath, WizardMath, and PromptCoT further evolve problems via model-driven generation~\cite{numina_math_7b,luo2023wizardmath,zhao2025promptcot}, while self-alignment and Magpie refine prompts with minimal supervision~\cite{li2023self,xu2024magpie}. Together, these approaches confirm LLMs’ ability to produce diverse, high-quality synthetic datasets for enhancing math reasoning. Our approach builds on this line of work by not only generating problems, but also explicitly increasing problem difficulty and structuring solution strategies to further enhance model reasoning capabilities.

\begin{figure*}[t]
\centering
\includegraphics[width=0.95\textwidth]{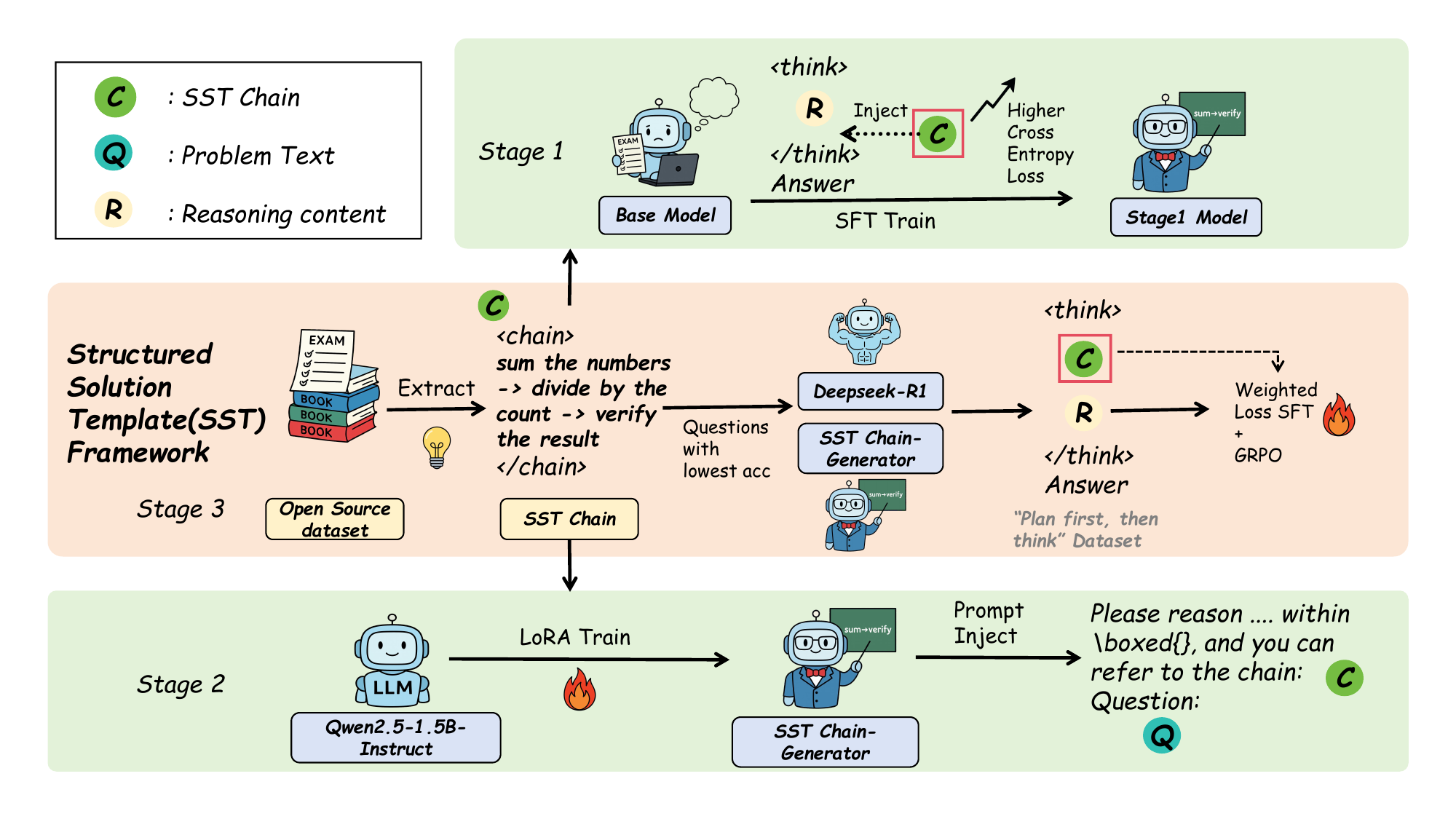}
\caption{
\textbf{Overview of our Structured Solution Template (SST) framework and three-stage training pipeline.}
\textbf{Bottom:} A lightweight LoRA-trained generator produces procedural SST chains, which are inserted at prompt time to guide solver models.
\textbf{Middle:} DeepSeek-R1 is used to generate structured ``Plan-then-Think'' solutions for challenging problems, creating an AI-generated curriculum of high-difficulty training data.
\textbf{Top:} These curated datasets are used to train the solver model with cross-entropy loss, assigning higher weight to SST chain tokens to emphasize procedural reasoning.
This framework encourages the model to internalize abstract solution strategies, reflecting the \textit{Scaling Law by Difficulty}: training on simple synthetic data results in fragile performance, while learning from carefully constructed, high-difficulty problems significantly enhances reasoning ability.
}
\label{fig:method_overview}
\end{figure*}

\noindent \textbf{Structured Chain-of-Thought}
Recent research also explores imposing explicit structure on the reasoning process~\cite{wei2022chain}. In code generation, Structured CoT (SCoT) separates logic planning from syntax generation, improving correctness~\cite{li2025structured}. Tabular CoT models reasoning as structured tables, clarifying multi-step solutions~\cite{jin2023tab}. More flexible systems like XoT combine multiple reasoning modes and verification steps, encouraging disciplined problem-solving~\cite{liu2023plan}. These methods suggest that guiding models with explicit reasoning templates enhances both reliability and generalization. Our method extends this line of work by systematically injecting structured solution strategies at multiple stages of training, further strengthening the model’s procedural reasoning abilities.

\noindent \textbf{Eliciting and Supervising Reasoning.} 
Providing intermediate steps—via prompting or training—substantially improves performance. CoT prompting enables models to solve complex tasks better than direct answering~\cite{wei2022chain}. Self-consistency samples multiple reasoning paths and aggregates results for robustness~\cite{wang2022self}. Tree-of-Thoughts explores reasoning branches via search, enabling deeper inference~\cite{yao2023tree}. Fine-tuning with rationales, as in MAmmoTH and WizardMath, further enhances math reasoning, with some models rivaling GPT-4~\cite{yue2023mammoth, luo2023wizardmath}. Methods like XoT integrate diverse reasoning strategies and iterative feedback, highlighting the benefits of supervising the reasoning process, not just final answers~\cite{liu2023plan}. Our approach is closely related to these methods, as we explicitly supervise intermediate reasoning steps and solution strategies to further strengthen model reasoning capabilities.

\section{Method}

\subsection{Scaling Law by Difficulty: Why More Data Can Hurt Reasoning?}

To address the shortcomings of existing approaches, we first investigate whether simply fine-tuning an LLM on large amounts of template-generated data can improve its reasoning on specific problem types. We start by selecting a complex ``parent'' problem with a detailed step-by-step solution and extract a general solution template (e.g., ``Step 1: Sum the numbers; Step 2: Divide by the count; Step 3: Verify the result''). Using an LLM as generator, we then create thousands of problem variations based on this template.

However, our results are surprising and counter-intuitive. As we increase the size of this synthetic dataset during fine-tuning, the model’s performance on held-out complex problems actually gets worse (see Fig.~\ref{fig:template_workflow}). Instead of becoming more robust at solving this type of problem, the model becomes less reliable.

\begin{figure}[htbp]  
    \includegraphics[width=\linewidth]{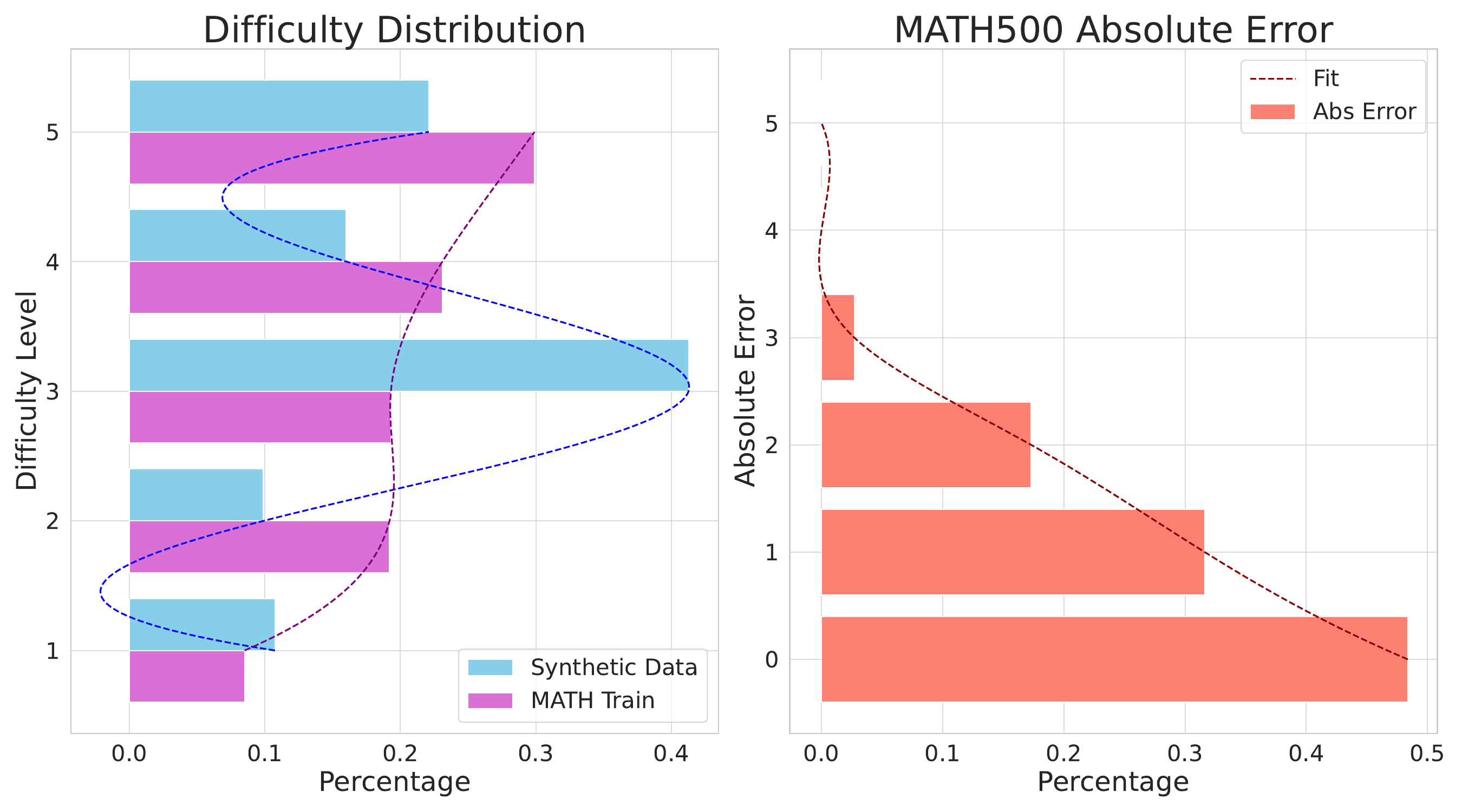}  
    \caption{
\textbf{Comparison of difficulty distribution between synthetic and MATH training data (left), and the absolute prediction error distribution on MATH500 test dataset (right). }
}
    \label{fig:template_workflow111}
\end{figure}

This unexpected outcome leads us to a new hypothesis: despite careful prompting, the generator model tends to produce cognitively simple problems that only mimic the surface features of the original, rather than its core reasoning steps. To test this, we sample 1000 ``parent'' problems from the MATH dataset, generate new problems with similar solution approaches using our template method, and ask GPT-4 to evaluate whether the generated problems match the original in difficulty (see Fig.~\ref{fig:template_workflow111}). The results confirm our suspicion: most generated problems are much easier than their originals. This highlights a key challenge---LLMs, when asked to generate difficult problems at scale, default to creating simpler versions, resulting in a curriculum that promotes shallow pattern matching rather than deep reasoning.

To provide a clear contrast, we  then conduct experiments on the Open-R1 dataset using only about one-fifth the data volume of the synthetic set, yet achieve a much higher score of 30.48 on the AIME24 benchmark. This demonstrates that carefully curated, high-difficulty problems are far more effective for improving model reasoning than large amounts of easy, synthetic data.

Taken together, these findings form the empirical foundation of what we call the \textbf{Scaling Law by Difficulty}:
\begin{itemize}
    \item Training on abundant, synthetic, low-difficulty data encourages shortcut heuristics and hurts performance.
    \item Training on carefully selected, high-difficulty data enforces true procedural reasoning and boosts performance.
\end{itemize}

In summary, our results show that advanced reasoning in LLMs depends not on the sheer volume of training data, but on the cognitive complexity of the problems. Overcoming shallow pattern matching requires deliberately focusing on hard, thoughtfully designed training examples that push models beyond memorized templates.

\subsection{Structured Soltion Template Framework}
Building on our findings from the previous section, we recognize that simply increasing data volume is not enough—in fact, it often reinforces the model’s reliance on surface-level patterns. The core idea behind our method is to enhance model reasoning by increasing the difficulty of training data through structured data augmentation. To this end, we introduce the SST framework (see Fig.~\ref{fig:method_overview}), which uses carefully designed templates to generate more challenging problems and make procedural solution steps explicit during training. This approach encourages the model to internalize abstract reasoning strategies rather than memorize shallow templates. In the following sections, we detail the three-stage design of SST.

\subsubsection{Stage 1: Fine-Tuning with Weighted Structured Solution-Template Chains}
The initial stage focuse on training the model to internalize not merely the solution steps, but also their underlying abstract procedures. We utilize the Open-R1 dataset, composed exclusively of challenging mathematical problems, deliberately chosen to encourage genuine procedural abstraction and reduce superficial pattern matching.

To augment this dataset, we apply a novel annotation method: an expert model (Qwen-32B Instruct) extract high-level summaries of logical procedures from gold-standard solutions with carefully crafted prompt. These summaries, termed \textit{structured solution-template}, encapsulate the abstract blueprint of reasoning and are appended in XML-style tags (\texttt{<chain>...</chain>}) at the end of each solution.

To guide the model toward prioritizing these procedural abstractions, we introduce a targeted weighting scheme during supervised fine-tuning (SFT). Specifically, we amplified the cross-entropy loss for tokens within the \texttt{<chain>} tags. The total loss function \$L\$ was defined as:
\begin{equation}
L = \frac{1}{N} \left( \sum_{i \notin \text{chain}} L_{CE}(y_i, \hat{y}_i) + w(t) \sum_{i \in \text{chain}} L_{CE}(y_i, \hat{y}_i) \right)
\label{eq:weighted-loss-global-norm}
\end{equation}
where $L_{CE}$ denotes the token-level cross-entropy loss, \$w(t)\$ represents a dynamically adjusted weight applied to the procedural chain tokens at training step $t$, and $N$ is the total number of tokens.

The weight $w(t)$ followed a linear decay schedule, beginning at an initial value $w_{initial}$ and gradually reducing to 1 over the training process:
\begin{equation}
w(t) = w_{initial} - (w_{initial} - 1) \frac{t}{T}
\label{eq:weight-decay}
\end{equation}
Here, $T$ denotes total training steps. This strategy emphasizes abstract reasoning early in training and transitions smoothly toward generalization by the end.

\subsubsection{Stage 2: Prompt-Time Chain Injection}
The second stage explored dynamic, prompt-time interventions. Here, the aim is to investigate whether a small, specialized model could enhance a larger model's reasoning by generating procedural scaffolds in real-time, effectively acting as a modular reasoning component.

We train a specialized chain-generator by fine-tuning a lightweight 1.5B-parameter model (Qwen-1.5B-Instruct) on structured solution-template from the Open-R1 dataset using Low-Rank Adaptation (LoRA)~\cite{hu2022lora}. During inference, this chain-generator produces concise, problem-specific solution templates. For example, given the prompt: ”\textbf{Problem:} In a right triangle, the lengths of the two legs are 3 cm and 4 cm. Find the length of the hypotenuse.“
The chain-generator outputs:
\begin{tcolorbox}[colback=white,colframe=black,boxrule=0.5pt,arc=2pt,left=2pt,right=2pt,top=2pt,bottom=2pt,width=\linewidth]
\small
\texttt{<chain>\\
Square each leg → \\Sum the squares →\\
Take the square root\\
</chain>}
\end{tcolorbox}

Building on this example, inference proceeds in two phases: first, the chain generator constructs a solution-template for the given problem; then, this chain is prepended to the original prompt and fed to the main solver, serving as an explicit cognitive scaffold to guide its reasoning.

This stage evaluates whether prompt-time injection of SST can replicate the benefits of high-difficulty training data, as identified by the Scaling Law by Difficulty—specifically, whether it can enhance the solver’s reasoning accuracy and efficiency without altering model weights, and potentially reduce computational overhead.

\subsubsection{Stage 3: Integrated Fine‐Tuning on an AI‐Generated Curriculum}
In this final stage, we merge the approaches of Stage 1 and Stage 2 into a single, unified curriculum that teaches the model to both plan and execute autonomously. We begin by rejection-sampling roughly 20,000 problems from the Open-R1 dataset on which the base model’s accuracy is lowest. For each sampled problem, we apply the Stage 2 chain-generator to propose a structured solution template and then invoke the DeepSeek-R1 API to produce a complete reasoning trace that (1) opens with the generated template, (2) follows each step in sequence, and (3) includes a self-reflection step allowing the model to revise its own chain mid-process. From the resulting ~10,000 examples, we distill the 3,843 instances exhibiting both high template fidelity and correct final answers into our core supervised set:

\begin{tcolorbox}[colback=white,colframe=black,boxrule=0.5pt,arc=2pt,left=2pt,right=2pt,top=2pt,bottom=2pt,width=\linewidth]
\small
\texttt{<think>\\
<chain>[Structured Solution Template Chain]</chain>[DETAILED STEPS]\\
…\boxed{ANSWER}}
\end{tcolorbox}

Unlike Stage 1, here the procedural chain appears at the very beginning of the trace, explicitly teaching the model to articulate a high-level plan before detailed execution. We first cold-start the base model via SFT on these 3,843 examples to instill the \texttt{<think>…<chain>…} format. We then perform GRPO reinforcement learning over the full 20,000-sample pool to optimize the model’s ability to generate accurate structured templates and to systematically follow and self-correct, according to them. We also apply the same weighted loss amplification for tokens within the \texttt{<chain>} tags as in Stage 1 to further emphasize SST. This integrated curriculum embeds robust “plan-then-execute” reasoning capabilities directly into the model, yielding a self-contained and reliable problem-solving system.

\section{Experiments}

\subsection{Datasets and Benchmarks}
We train our models on the OpenR1 dataset (default 94k-sample split from OpenR1-Math-220k), a high-quality collection of math problems with multi-step solution traces~\cite{openr1}. For evaluation, we employ a diverse set of reasoning benchmarks: AIME 2024/2025~\cite{aime2024,aime2025}, AMC 2023~\cite{amc12a2023}, GSM8K~\cite{cobbe2021gsm8k}, MATH500~\cite{lightman2023lets}, and Gaokao English 2023~\cite{gaokao2023_math_en}. We further introduce \emph{Dynamic Math}, a new interdisciplinary benchmark covering nonlinear dynamics in physics, chemistry, and mathematics, designed to assess cross-domain generalization.

\subsection{Model Setup and Training}
Our primary model is DeepSeek-R1-Distill-Qwen-1.5B, a Qwen2.5-derived model pre-trained on 800K reasoning examples~\cite{guo2025deepseek}. Reasoning templates are extracted using Qwen2.5-14B-Instruct, while a lightweight chain generator is trained by fine-tuning Qwen2.5-Math-1.5B-Instruct with LoRA on the extracted templates~\cite{qwen2.5}. In Stage 3, we utilize the DeepSeek-R1 API along with the Stage 2 chain generator to construct an initial structured dataset of 3k examples, followed by mining an additional 20k problems for GRPO training. The complete pipeline integrates: \emph{DeepSeek-R1-32B} (generation and verification), \emph{Qwen2.5-14B-Instruct} (chain extraction), and \emph{Qwen2.5-1.5B-Instruct (LoRA)} (chain generation).

\subsection{Compared Baselines}
We conduct a comprehensive comparison between SST and a range of recent math reasoning models at the 1.5B and 7B parameter scales. Specifically, our baselines include Qwen2.5-Math (both base and instruct variants), DeepSeek-R1-Distill (1.5B and 7B)\cite{guo2025deepseek}, OpenRS1, OpenRS2, and OpenRS3\cite{dang2025reinforcement}, PromptCoT-DS-1.5B~\cite{zhao2025promptcot}, as well as L1-Qwen-1.5B (Max and Exact)~\cite{aggarwal2025l1}. With the exception of PromptCoT, all baseline models are fine-tuned from either Qwen2.5 or DeepSeek-R1 using reinforcement learning techniques. For fair evaluation, we use the publicly released model weights and apply identical inference configurations across all models. This ensures that our comparative results reflect differences in model capabilities and training strategies, rather than discrepancies in evaluation protocol.

\subsection{Evaluation Details}
We report accuracy based on the final boxed numeric answer, averaged over eight random seeds with results shown as \textit{mean~$\pm$~std} (e.g., 44.0~$\pm$~4.5). No cherry-picking or post-tuning is performed. Inference follows each model's default configuration: most use temperature 0.6 and top-$p$ 0.95; Qwen models use temperature 0 and top-$p$ 1.0. For Stage 2 chain generation, we set temperature to 0.4 and top-$p$ to 1.0 to improve consistency. Output is capped at 16,384 tokens. All experiments are performed on NVIDIA A100 40GB GPUs.

\subsection{Main Results}

\begin{table*}[t]
\centering
\caption{\textbf{Main results (accuracy $\pm$ std) across seven reasoning benchmarks.} Step 1 uses weighted chain-supervised fine-tuning; Step 2 applies prompt-time procedural chain injection; Step 3 refers to integrated curriculum finetuning. Only the top-3 scores in each column are shaded. Top-3 models in each category are
highlighted with increasing gray intensity}
\label{tab:main-results}
\resizebox{\textwidth}{!}{
\begin{tabular}{lccccccc}
\toprule
\textbf{Model}                  & \textbf{GSM8K}                & \textbf{MATH500}              & \textbf{AIME24}               & \textbf{AIME25}               & \textbf{Gaokao En}            & \textbf{AMC23}                & \textbf{Dynamic En}           \\
\midrule
Qwen-1.5B-Base                  & 59.24 ± 0.30                  & 35.60 ± 0.94                  &  1.67 ± 1.67                  &  0.00 ± 0.00                  & 30.81 ± 1.14                  & 23.98 ± 1.04                  & 12.50 ± 1.77                  \\
Qwen-MATH-1.5B                  & 25.82 ± 0.31                  & 18.95 ± 0.61                  &  0.00 ± 0.00                  &  0.00 ± 0.00                  & 14.10 ± 0.18                  &  3.20 ± 0.13                  &  8.75 ± 2.80                  \\
Qwen-MATH-1.5B-Instruct         & \toptwo{83.85 ± 0.20}         & 75.00 ± 0.65                  &  8.33 ± 1.67                  & 10.00 ± 0.00                  & 67.23 ± 0.68                  & 63.87 ± 1.33                  & 52.12 ± 2.07                  \\
Qwen2.5-Math-1.5B-Oat-Zero      & 82.22 ± 0.27                  & 74.60 ± 1.67                  & 17.50 ± 2.76                  & 15.00 ± 2.89                  & 61.29 ± 1.05                  & 27.88 ± 0.57                  & 51.88 ± 6.47                  \\
\midrule
DeepSeek-1.5B-Distill           & 78.62 ± 0.58                  & 83.10 ± 0.72                  & \toptwo{28.30 ± 5.88}         & 22.67 ± 5.34                  & 72.89 ± 0.95                  & 69.00 ± 5.85                  & 54.19 ± 1.09                  \\
OpenRS1                         & 80.25 ± 0.97                  & 84.40 ± 1.24                  & 27.92 ± 4.39                  & 23.33 ± 5.77                  & 73.79 ± 1.29                  & 70.00 ± 5.00                  & 54.86 ± 2.63                  \\
OpenRS2                         & 79.70 ± 0.41                  & \topthree{84.63 ± 1.33}       & \topthree{28.11 ± 4.16}       & 23.33 ± 4.71                  & 73.50 ± 0.78                  & 68.75 ± 7.03                  & 54.79 ± 1.62                  \\
OpenRS3                         & 79.12 ± 0.65                  & \toptwo{84.67 ± 1.18}         & 26.67 ± 3.33                  & 21.11 ± 3.69                  & 72.80 ± 1.09                  & 67.50 ± 2.04                  & 54.31 ± 2.02      \\
L1-Qwen-1.5B-Exact              & 82.10 ± 0.32                  & \topone{86.20 ± 0.82}         & 23.75 ± 4.55                  & \topthree{24.17 ± 3.23}       & \topone{78.52 ± 0.76}         & \toptwo{73.12 ± 5.27}         & \topone{63.73 ± 1.89}         \\
L1-Qwen-1.5B-Max                & \topthree{82.44 ± 0.45}       & 84.03 ± 0.54                  & 25.42 ± 4.39                  & \topone{26.25 ± 3.89}         & \toptwo{77.97 ± 0.54}         & \topone{75.31 ± 4.58}         & \toptwo{62.40 ± 1.41}         \\
Prompt-COT-DS-1.5B              & 81.14 ± 0.66                  & 77.47 ± 1.07                  & 21.88 ± 7.81                  & 21.25 ± 5.12                  & 68.68 ± 1.63                  & 61.25 ± 6.43                  & 38.59 ± 1.67                  \\
\midrule
SST-Framework (Ours)            & \topone{84.80 ± 0.69}         & 84.15 ± 0.94                  & \topone{30.48 ± 2.60}         & \toptwo{25.00 ± 5.53}         & \topthree{74.93 ± 1.48}       & \topthree{71.88 ± 8.64}       & \topthree{55.00 ± 3.92}                  \\
\bottomrule
\end{tabular}
}
\end{table*}

A comprehensive comparison between our proposed method and the state-of-the-art baselines is presented in Table~\ref{tab:main-results}. From the quantitative results, we identify three important conclusions, which are discussed in detail as follows.

\noindent \textbf{Substantial end-to-end performance improvements from the SST framework.}  
    Our SST-Framework achieves top-tier accuracy on all seven benchmarks, often surpassing both base models and prior state-of-the-art. For instance, on GSM8K we record 84.80 $\pm$ 0.69—a 6.18-point absolute gain over DeepSeek-1.5B-Distill  (78.62 ± 0.58) and a 2.36-point lead over the previous best single-model result (82.44 $\pm$ 0.45 by L1-Qwen-1.5B-Max). Similarly, on AIME24 we attain 30.48 $\pm$ 2.60, compared to 28.30 ± 5.88 for the base and 23.75 $\pm$ 4.55 for L1-Qwen-1.5B-Exact—an 2.18-point boost. Even on already strong benchmarks such as MATH500 (84.15 $\pm$ 0.94) and AMC23 (71.88 $\pm$ 8.64), SST-Framework remains within the top three, demonstrating that weighted chain-supervised fine-tuning, prompt-time chain injection, and integrated curriculum fine-tuning collectively elevate LLM reasoning across the board.

\noindent \textbf{Indirect validation of the Scaling Law by Difficulty via prompt-time chain injection.}  
Our results also corroborate the Scaling Law by Difficulty by showing that structured chain prompts confer the largest benefits on lower-difficulty tasks while still delivering measurable gains on harder ones. For example, Gaokao English improves from 72.89 $\pm$ 0.95 to 74.93 $\pm$ 1.48 (+2.05) and GSM8K from 78.62 $\pm$ 0.58 to 84.80 $\pm$ 0.69 (+6.18), whereas AIME25 rises from 22.67 $\pm$ 5.34 to 25.00 $\pm$ 5.53 and MATH500 from 83.10 $\pm$ 0.72 to 84.15 $\pm$ 0.94. This pattern highlights that our structured solution template chains scaffold model reasoning in a difficulty-adaptive manner—yielding larger boosts on simpler benchmarks and consistent improvements on more challenging ones.

\noindent \textbf{Dynamic English (“Dynamic En”) as a high-discriminative, cross-domain benchmark.}  
    The Dynamic En dataset integrates advanced problems spanning physics, chemistry, and mathematics, producing a wide performance spectrum—from 8.75 ± 2.80 (Qwen-MATH-1.5B) and 12.50 ± 1.77 (Qwen-1.5B-Base) up to 63.73 ± 1.89 (L1-Qwen-1.5B-Exact).  Common chain-based baselines cluster in the mid-50s: Qwen-MATH-1.5B-Instruct at 52.12 ± 2.07, DeepSeek-1.5B-Distill at 54.19 ± 1.09, OpenRS1 at 54.86 ± 2.63, and OpenRS2 at 54.79 ± 1.62.  Our SST-Framework achieves 55.00 ± 3.92, outperforming most prior methods and demonstrating that Dynamic En sharply distinguishes between weak, moderate, and strong reasoning models—making it an exceptionally rigorous benchmark for cross-domain procedural reasoning.

\begin{table*}[t]
\centering
\caption{\textbf{Performance of the three stages across seven benchmarks (accuracy $\pm$ std).}}
\label{tab:step-comparison}
\resizebox{\textwidth}{!}{
\begin{tabular}{lccccccc}
\toprule
\textbf{Stage}                                       & \textbf{GSM8K}            & \textbf{MATH500}         & \textbf{AIME24}          & \textbf{AIME25}          & \textbf{Gaokao En}       & \textbf{AMC23}           & \textbf{Dynamic En}      \\
\midrule
Stage 1 (Chain-Weighted SFT)                        & 80.18 ± 0.61              & \textbf{84.83 ± 0.99}             & 27.71 ± 5.98             & 23.75 ± 3.89             & 73.99 ± 1.43             & 70.53 ± 3.45    & 53.66 ± 1.39  \\
Stage 2 (Prompt-Time Procedural Chain Injection)     & 73.79 ± 0.54              & 82.56 ± 1.13             & \textbf{32.67 ± 3.89}    & \textbf{27.33 ± 5.33}    & 68.88 ± 0.69             & 71.90 ± 4.06             & 54.26 ± 2.29             \\
Stage 3 (SST) & \textbf{84.80 ± 0.69}     & 84.15 ± 0.94             & 30.48 ± 2.60             & 25.00 ± 5.53             & \textbf{74.93 ± 1.48}             & \textbf{71.88 ± 8.64}             & \textbf{55.00 ± 3.92}             \\
\bottomrule
\end{tabular}
}
\end{table*}

\begin{table}[H]
\centering
\caption{\textbf{Impact of Prompt-time Chain Injection on Accuracy and Token Efficiency.} Accuracy (left) and average output token count (right) per problem during Step 2 inference, comparing runs without and with prompt-injected chains. The percentage shows the relative change in token count after chain injection (negative indicates reduction, positive indicates increase).}
\label{tab:token-efficiency}
\resizebox{\columnwidth}{!}{
\begin{tabular}{lcccc}
\toprule
\textbf{Dataset} & \textbf{w/o chain} & \textbf{with chain} & \textbf{Tokens Reduction (\%)} \\
\midrule
GSM8K &\textbf{78.62}/1249.89 & 73.79/469.89& \textbf{-62.4\%}\\
AIME24 &28.30/12478.98 & \textbf{32.67}/11994.29& \textbf{-3.9\%}\\
AIME25 &22.67/11924.05 & \textbf{27.33}/11970.28& +0.3\%\\
AMC23 &69.00/7984.44 & \textbf{71.90}/7673.46& \textbf{-3.9\%}\\
\bottomrule
\end{tabular}
}

\end{table}

\noindent \subsection{Further Analysis}

Building on our main results, we proceed to: (1) isolate the contribution of each stage across multiple benchmarks; (2) evaluate the impact of prompt-time chain injection on both efficiency and solution conciseness; and (3) investigate the interaction between weighted chain training and dynamic prompting. Below is the detailed analysis.

\paragraph{Stage-wise Contributions (Table \ref{tab:step-comparison}).}  
Stage 1’s chain-weighted supervised fine-tuning already yields strong procedural priors: it achieves the highest accuracy on MATH500 (84.83 ± 0.99) and competitive performance on Gaokao En (73.99 ± 1.43) and AMC23 (70.53 ± 3.45), indicating that emphasizing abstract solution templates during training substantially improves performance on difficult, curated problems. However, Stage 1 underperforms on problems requiring real-time scaffolding (AIME24/25), where dynamic guidance is crucial. Stage 2’s prompt-time procedural chain injection flips this pattern: it attains the best results on AIME24 (32.67 ± 3.89) and AIME25 (27.33 ± 5.33), but sacrifices gains on MATH500 and GSM8K, suggesting that on-the-fly chains are most effective for moderate‐difficulty puzzles but insufficient alone for the hardest benchmarks. Stage 3 (SST) integrates both mechanisms and consistently outperforms or matches the best of Stages 1–2 across all tasks—e.g., GSM8K rises to 84.80 ± 0.69 and Gaokao En to 74.93 ± 1.48—demonstrating complementary benefits of weighted template learning and dynamic chain prompting.

\paragraph{Inference Efficiency and Difficulty Scaling (Table \ref{tab:token-efficiency}).}  
When comparing Step 2 runs with and without prompt-injected chains, we observe a difficulty-dependent trade-off between accuracy and output length, echoing the Scaling Law by Difficulty. On simpler benchmarks like GSM8K, accuracy slightly decreases (78.62 → 73.79) while token count plummets by 62.4 \% (1,249.89 → 469.89), suggesting that the model “blindly” follows the injected chain with minimal extra reasoning. In contrast, on harder tasks such as AIME24 and AIME25, accuracy actually improves (AIME24: 28.30 → 32.67; AIME25: 22.67 → 27.33) but token reduction is modest (–3.9 \% and +0.3 \%, respectively), indicating the model engages in more reflective thought rather than passively adhering to the chain. AMC23 likewise shows a 3.9 \% reduction in tokens alongside a 2.87-point accuracy gain. These results imply that for low-difficulty problems, prompt-injected chains streamline and truncate reasoning traces, whereas for high-difficulty problems they serve as a scaffold that still prompts active, deliberative reasoning.

\paragraph{Synergy of Components.}  
Together, these analyses underline that (1) weighted chain supervision grounds the model in abstract procedures, (2) prompt-time chains dynamically scaffold reasoning where it matters most, and (3) their integrated application (Stage 3) yields robust, efficient performance across simple to highly complex tasks. This synergy validates our design choices and highlights pathways for further optimization in structured reasoning frameworks.  

\subsection{Ablation Study}

\begin{table}[t]
\centering
\caption{\textbf{Ablation Study on Stage 1.} Impact of chain annotation and weighted loss on GSM8K and Dynamic English (accuracy).}
\label{tab:chain-annotation}
\resizebox{\columnwidth}{!}{
\begin{tabular}{lcc}
\toprule
\textbf{Training Method} & \textbf{GSM8K} & \textbf{Dynamic En} \\
\midrule
Stage 1 Without \texttt{<chain>} & \textbf{81.88}&52.00 \\
Stage 1 With \texttt{<chain>}, w/o weighted loss & 80.36&53.35  \\
Stage 1 With \texttt{<chain>}, with weighted loss & \underline{80.88}&\textbf{53.66} \\
\bottomrule
\end{tabular}
}
\end{table}

\begin{figure}[!b]
  \centering
  \includegraphics[width=\columnwidth]{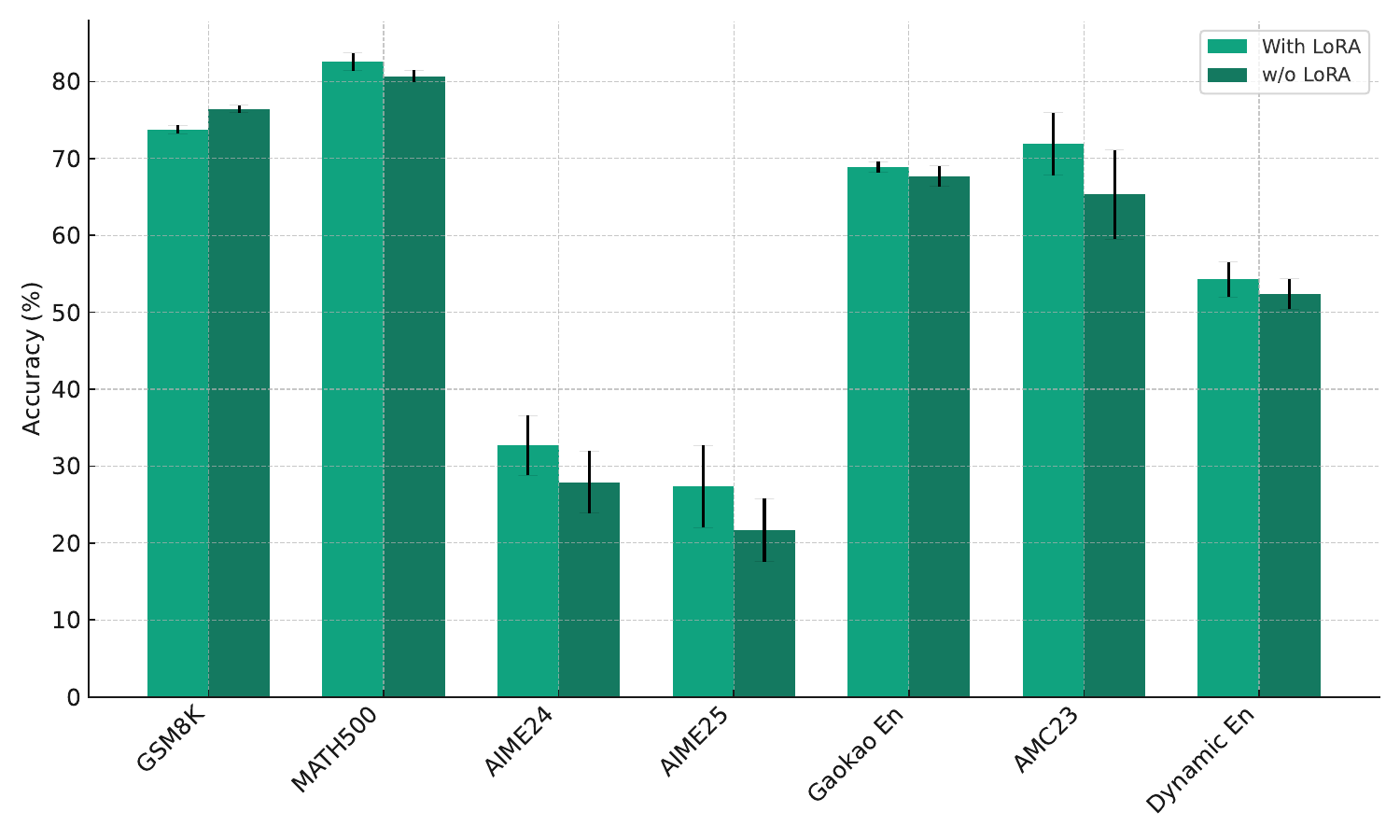}
  \caption{\textbf{Comparison of chain‐generator models in Stage 2}: pass@1 accuracy (mean$\pm$std) on GSM8K, AIME24 and AMC23, for model with and without LoRA.}
  \label{fig:chain-generator}
\end{figure}

\noindent \textbf{Impact of the structured chain annotation and weighted loss (Table \ref{tab:chain-annotation}).}  
    Simply adding \texttt{<chain>} without weighted loss introduces a procedural scaffold but can bias the model toward surface templates: GSM8K drops from 81.88 to 80.36, while Dynamic En rises from 52.00 to 53.35. Applying our dynamically decayed weighted loss counteracts this bias, recovering GSM8K to 80.88 and further boosting Dynamic En to 53.66. This demonstrates that emphasizing abstract chain tokens during SFT is crucial to leverage template annotation without overfitting to shallow heuristics.

\begin{table}[H]
\centering
\caption{\textbf{Ablation Study on Stage 3.} Effect of weighted loss during SST cold start on GSM8K and Dynamic English.}
\label{tab:weighted-loss-ablation}
\resizebox{\columnwidth}{!}{
\begin{tabular}{lcccc}
\toprule
\textbf{Training Method} & \textbf{GSM8K} & \textbf{Dynamic En} \\
\midrule
SST cold start, w/o weighted loss & 82.16&42.20  \\
SST cold start with weighted loss & 82.61&48.28 \\
\bottomrule
\end{tabular}
}
\end{table}

\noindent \textbf{Impact of the weighted soss vs. unweighted (Table \ref{tab:weighted-loss-ablation}).}  
    Applying weighted loss to the chain tokens recovers GSM8K accuracy to 80.88 (+0.52) and further improves Dynamic En to 53.66 (+0.31). This confirms that emphasizing procedural chains in the loss function helps the model better leverage abstract templates and mitigates performance drops on simpler benchmarks.

\noindent \textbf{Impact of LoRA on the chain generator. (Fig. \ref{fig:chain-generator})}  
    Without LoRA, the chain generator achieves 76.44 ± 0.49 on GSM8K, 80.70 ± 0.77 on MATH500, and 52.37 ± 1.97 on Dynamic En.  
With LoRA, it obtains 73.79 ± 0.54 on GSM8K, 82.56 ± 1.13 on MATH500, and 54.26 ± 2.29 on Dynamic En.  
Incorporating LoRA yields a 1.86-point gain on MATH500 and a 1.89-point gain on Dynamic En, at the cost of a 2.65-point drop on the simpler GSM8K. This suggests that parameter-efficient adaptation enhances scaffold fidelity on more challenging cross-domain tasks, while slightly reducing chain quality for easier problems.  

\noindent \textbf{Impact of the role of GRPO in curriculum fine-tuning (Table \ref{tab:cold-start}).}  
    In the SST cold-start setting, enabling GRPO boosts GSM8K by +2.19 (82.61 → 84.80), MATH500 by +4.35 (79.80 → 84.15), and Dynamic En by +6.72 (48.28 → 55.00). These substantial gains confirm that a hard-mined, difficulty-balanced curriculum with guided reasoning plan optimization is instrumental in driving the model beyond superficial patterns toward genuine procedural abstraction.

\begin{table}[H]
\centering
\caption{\textbf{Ablation Study on w/ vs. w/o Cold Start} Effect of GRPO on cold-start SST performance (accuracy) across GSM8K, MATH500, and Dynamic English.}
\label{tab:cold-start}
\resizebox{\columnwidth}{!}{
\begin{tabular}{lccc}
\toprule
 & \textbf{GSM8K}& \textbf{MATH500} & \textbf{Dynamic En} \\
\midrule
SST w/o GRPO & 82.61& 79.80& 48.28 \\
SST with GRPO & \textbf{84.80}& \textbf{84.15}& \textbf{55.00} \\
\bottomrule
\end{tabular}
}
\end{table}

\section{Conclusion}

This paper uncovers the Scaling Law by Difficulty, revealing that model performance on math reasoning tasks follows a U-shaped curve with respect to training data complexity—highlighting the limitations of excessive low-difficulty data and the benefits of high-difficulty examples. Building on this insight, we propose the Structured Solution Template framework, which combines structured solution templates and a difficulty-aware curriculum to explicitly teach procedural reasoning to LLMs. Experimental results on GSM8K, AIME24, and our Dynamic Math benchmark demonstrate that SST significantly enhances both reasoning accuracy and efficiency, particularly on challenging problems.  

\noindent \textbf{Limitation.} The current framework is primarily evaluated on mathematics, and its generalization to other complex reasoning tasks warrants further exploration.

\bibliography{aaai2026}

\end{document}